\documentclass{article} 
\usepackage{iclr2027_conference,times}

\usepackage{amsmath,amsfonts,bm}

\def\eqref#1{equation~\ref{#1}}

\def\1{\bm{1}}

\def\ra{{\textnormal{a}}}

\def\rx{{\textnormal{x}}}

\def\rva{{\mathbf{a}}}

\def\erva{{\textnormal{a}}}

\def\ervx{{\textnormal{x}}}

\def\rmA{{\mathbf{A}}}

\def\vmu{{\bm{\mu}}}
\def\vtheta{{\bm{\theta}}}
\def\va{{\bm{a}}}

\def\ve{{\bm{e}}}

\def\vx{{\bm{x}}}

\def\eva{{a}}

\def\mA{{\bm{A}}}

\def\mH{{\bm{H}}}
\def\mI{{\bm{I}}}
\def\mJ{{\bm{J}}}

\def\mX{{\bm{X}}}

\def\mSigma{{\bm{\Sigma}}}

\DeclareMathAlphabet{\mathsfit}{\encodingdefault}{\sfdefault}{m}{sl}
\SetMathAlphabet{\mathsfit}{bold}{\encodingdefault}{\sfdefault}{bx}{n}
\newcommand{\tens}[1]{\bm{\mathsfit{#1}}}
\def\tA{{\tens{A}}}

\def\tX{{\tens{X}}}

\def\gG{{\mathcal{G}}}

\def\sA{{\mathbb{A}}}
\def\sB{{\mathbb{B}}}

\def\sS{{\mathbb{S}}}

\def\emA{{A}}

\newcommand{\etens}[1]{\mathsfit{#1}}

\def\etA{{\etens{A}}}

\newcommand{\E}{\mathbb{E}}

\newcommand{\R}{\mathbb{R}}

\newcommand{\KL}{D_{\mathrm{KL}}}
\newcommand{\Var}{\mathrm{Var}}

\newcommand{\Cov}{\mathrm{Cov}}

\newcommand{\normltwo}{L^2}
\newcommand{\normlp}{L^p}

\newcommand{\parents}{Pa} 

\usepackage{hyperref}
\usepackage{url}
\usepackage{graphicx}
\usepackage{booktabs}
\usepackage{multirow}
\usepackage{wrapfig}
\usepackage{anyfontsize}

\newcommand{\tabfont}{\fontsize{7.5}{9}\selectfont}

\title{Dynamic Expert Pruning for Multi-Agent \\ Systems}

\author{Jabin Koo$^{1}$ \quad Soheil Abbasloo$^{2}$ \quad Sungjae Lee$^{1}$ \quad Jungseul Ok$^{1}$ \\
$^{1}$Pohang University of Science and Technology (POSTECH),
$^{2}$Microsoft Research Asia \\
}

\iclrfinalcopy 
\begin{document}

\maketitle
\ificlrfinal\lhead{}\fi

\begin{abstract}
Mixture-of-Experts (MoE) architectures scale language models efficiently by
activating only a few experts per token, but the saving is confined to
computation: every expert must stay resident on the accelerator, so memory
bounds where these models can be deployed. Expert pruning reduces this
footprint, yet existing methods are static --- a single mask, calibrated
offline, is applied to the model for every subsequent request. This
assumption can fail when the workload is heterogeneous, most prominently in
multi-agent systems,
where one backbone serves many tasks and roles at once: our analysis shows
that different tasks and roles recruit different experts, while static
methods assign one fixed subset to all of them. We therefore propose
\emph{Dynamic Expert Pruning} (DEP), which rests on a finding we establish
here: an agent's system and task prompts are by themselves sufficient to
identify the experts that agent and its task require, since that text already
describes what the agent will do. A lightweight predictor, trained once on workflow
transcripts, turns those prompts into a specialized per-request mask in a
single forward pass, with no per-configuration calibration. Across diverse
tasks and roles, model scales, and MoE architectures, DEP achieves
better overall accuracy than static pruning and merging baselines, and
generalizes to workflows unseen in training without retraining. Its margin
over those baselines is largest when few experts are retained, suggesting
that the role specialization inherent to multi-agent systems permits sparser
serving than static pruning allows.

\end{abstract}

\section{Introduction}
\label{sec:intro}

Scaling language models has increasingly meant scaling them sparsely rather
than densely. A Mixture-of-Experts (MoE) layer replaces a single feed-forward
network with a pool of expert networks and a router that activates only a few
of them per token, so the parameter count can grow without a proportional
growth in per-token computation \citep{shazeer2017moe, fedus2022switch}. The
design now underpins the strongest open models \citep{jiang2024mixtral,
deepseekai2024v3, yang2025qwen3} and frontier proprietary systems alike
\citep{geminiteam2024gemini}. The saving it delivers, however, is in
arithmetic, not memory: routing is re-decided at every token, so any expert
may be needed at any step and the entire pool must stay resident on the
accelerator --- or be paged in at a latency cost. Memory, not compute,
therefore often determines where an MoE model can be deployed. This
asymmetry has made expert pruning an active line of work --- removing the
experts that contribute least \citep{lu2024naee, lasby2025reap},
or merging redundant ones \citep{chen2024hcsmoe} --- with a shared,
encouraging finding: a large fraction of the experts can be discarded at
modest loss, \emph{provided} the retained subset matches the workload the
model will actually serve.

Current methods simply assume that condition holds. Previous expert pruning
methods are calibration-based: expert importance is averaged over a corpus
taken to stand for future traffic, and the resulting \emph{static} mask is
frozen for every subsequent request. But a single backbone increasingly serves many
prompting contexts at once --- most prominently multi-agent systems, which
assign it different tasks and agentic roles through prompts alone. Static
calibration answers this diversity by finding one uniform, best-on-average
set of experts; we argue that different tasks and roles may require
different experts, so a single averaged subset ends up fitted to none of
them.

Our analysis bears this out. Static pruning methods that deliver on
single-workload benchmarks lose much of their accuracy in settings that mix
tasks and roles, multi-agent systems chief among them, and the loss has a
mechanism: agents in different roles and on different tasks recruit
measurably different experts, so preserving quality calls for an expert mask
\emph{specialized} to each agent and task. Specialization by calibration,
however, scales poorly --- one pass per configuration, redone whenever a
prompt is edited, a role is added, or a workflow is re-searched. This
motivates expert pruning that is \emph{dynamic}: specialized to each
request, without a calibration pass for each.

This is precisely what we propose: \textbf{Dynamic Expert Pruning (DEP)}.
Instead of calibrating a mask, DEP predicts it, with a lightweight predictor
that learns which experts an agent on a given task will need from nothing but
the agent's system prompt and its task input. A central claim of this paper
is that \emph{these prompts are by themselves sufficient}: the system prompt
specifies what the agent is and what it will do, and the task input specifies
what knowledge doing it demands, so the text that configures an agent is
enough on its own to identify its specialized expert set before a single
token is generated --- and our experiments bear this out, on agents and
workflows the predictor never saw. The predictor is trained on
transcripts of the backbone at work --- supervised by which experts
contributed to each output, and by how little the masked model's distribution
moves --- and at serving time the mask is held fixed for the turn, preserving
what makes static pruning deployable: a bounded resident expert set, reuse
across a batch, untouched weights. Because it conditions on prompt text
rather than a roster identifier, even agents and workflows never seen in
training receive their masks \emph{in one shot} --- a single small forward
pass on the system and task prompts, with no additional calibration.

Empirically, DEP achieves better overall accuracy than static pruning and
merging baselines across benchmarks, workflows, model scales, and MoE
architectures, and it generalizes to configurations unseen in training; our
analyses trace why. No single
static criterion is uniformly worse --- each wins somewhere --- but each
also collapses somewhere, and DEP is the only policy that does not. Its
advantage is largest in the
sparsest regime: as the budget tightens, a fixed subset must cover every
context and ends up serving none well, while DEP spends the whole budget on
the experts one request needs. These results suggest that specialization
opens the door to sparser pruning --- narrowing a mask to a single agent's
task and knowledge can let far fewer experts suffice.

\paragraph{Contributions.}
\begin{itemize}
\itemsep2pt
\item We analyze why static expert masks underperform in multi-agent
systems --- agents in different roles and on different tasks recruit
different experts, so any single mask is fitted to none of them ---
motivating dynamic, per-agent expert masks.
\item We show that an agent's system and task prompts are sufficient to
identify the experts its task and role require, and build DEP on that
finding: a mask predictor that reads only those prompts and emits a
specialized expert mask on demand, \emph{in one shot}, with no
per-configuration calibration.
\item We validate DEP across six agentic workflows, unseen structures,
families, benchmarks, and domains, on SWE-bench Verified at 30B and 480B
scale, and on a different MoE architecture, where it achieves the best overall
accuracy among pruning policies and separates from them by a large margin in
the sparsest regime --- evidence that per-agent specialization can enable
sparser serving.
\end{itemize}

\section{Background and Motivation}
\label{sec:motivation}

\begin{figure}[t]
\centering
\includegraphics[height=0.30\linewidth]{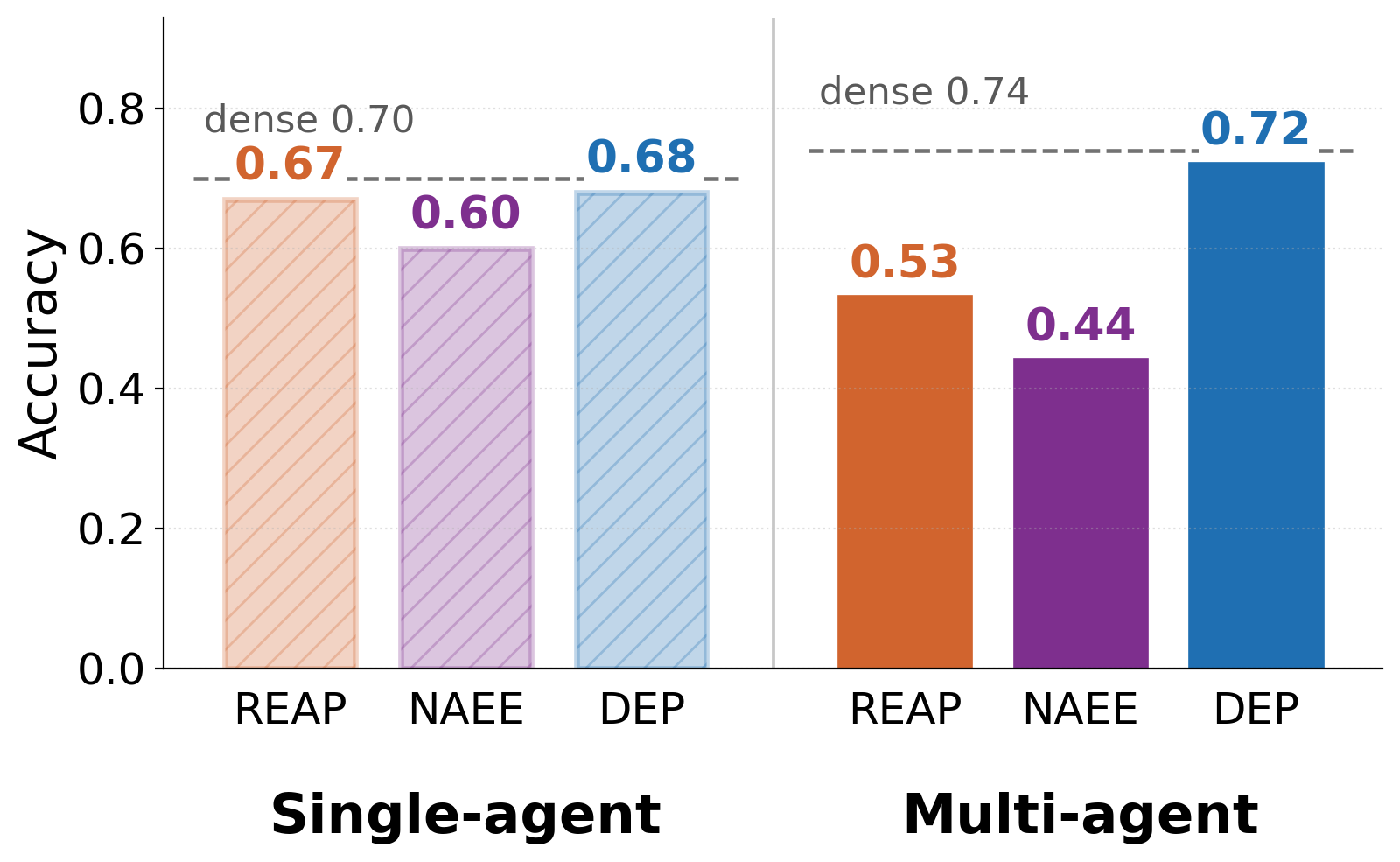}%
\hspace{0.02\linewidth}%
\includegraphics[height=0.30\linewidth]{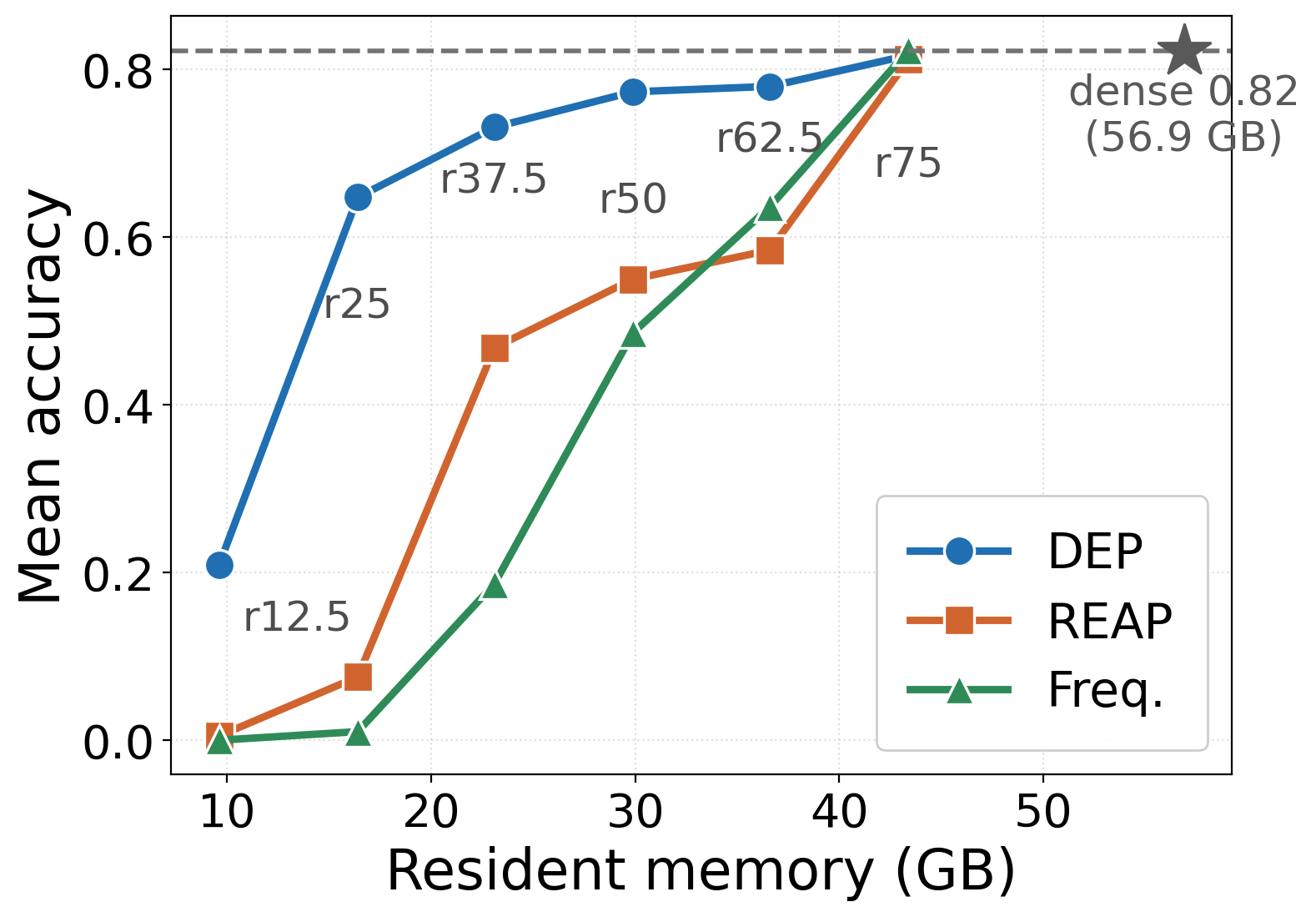}
\caption{\textbf{Static masks that track the dense model for a single agent
lose much of that accuracy on a multi-agent workload.}
\textbf{Left:} identical pruning methods, backbone, retention
($\rho{=}50\%$), and calibration data; only the serving setting changes.
\textbf{Right:} the resulting accuracy--memory frontier.}
\label{fig:single-vs-multi}
\end{figure}

Static expert pruning scores each expert on a calibration corpus ---
by routing frequency, reconstruction error \citep{lu2024naee}, or
router-weighted activation magnitude \citep{lasby2025reap} --- and freezes
the resulting subset, or a merged version of it \citep{chen2024hcsmoe}, for
every future request; the implicit assumption is that the corpus represents
the traffic. Figure~\ref{fig:single-vs-multi} shows what happens when it
does not. With the backbone, retention, and calibration data held fixed, the
same static masks that track the dense model when serving a single agent
lose a large fraction of its accuracy on a multi-agent workflow over the
same tasks, while a mask chosen per agent does not. The reversal matters
because the workflow is there to buy accuracy in the first place: unpruned,
the multi-agent system scores above the single agent, yet under the same
static masks it ends up \emph{below} it --- static pruning does not merely
cost accuracy here, it cancels the advantage the workflow was adopted to
gain. The methods did not get
worse; the workload changed underneath them --- one subset must now serve
many different roles and tasks at once, and it is fitted to none of them --- and the
failure is ungraceful: routing through the wrong experts degenerates into
repetition rather than merely answering worse
(Section~\ref{sec:efficiency}). Whether multi-agent compositions outperform
a single well-prompted model is actively debated \citep{cemri2025mast,
kim2025scaling, jwalapuram2026illusion}; nothing here depends on the answer,
only on the observation that requests reaching one backbone now carry
many different roles and instructions.

\begin{figure}[t]
\centering
\includegraphics[width=\linewidth]{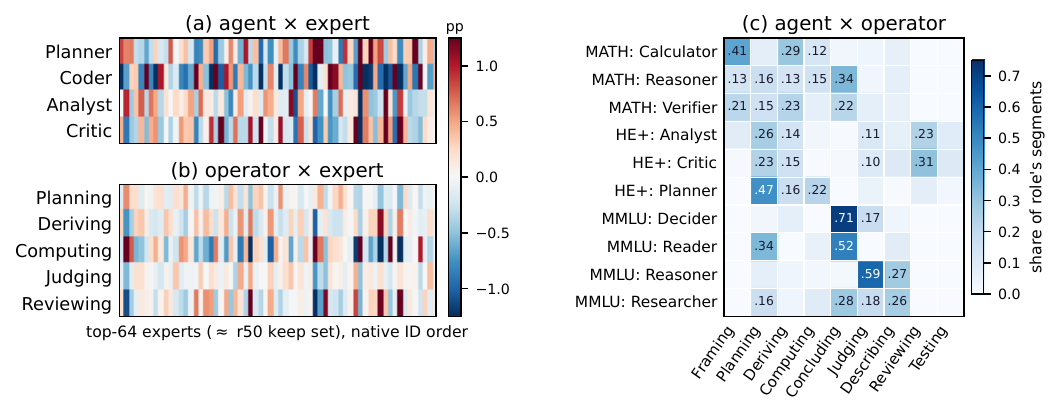}
\caption{\textbf{Agents differ in what they do, and therefore in the experts
they use.} \textbf{(a)} Per-role and \textbf{(b)} per-operator expert usage
(HumanEval+, layer 24; deviation from the row mean over the top-64 experts,
percentage points); \textbf{(c)} share of each role's segments per operator, across three tasks.}
\label{fig:motivation}
\end{figure}

Figure~\ref{fig:motivation} shows why. To describe agent behavior
independently of agent identity, we segment agent outputs into nine
\emph{discourse operators} --- Framing, Planning, Deriving, Computing,
Concluding, Judging, Describing, Reviewing, and Testing --- using a
segmentation methodology adapted from \citet{reasonops2026}
(Appendix~\ref{app:operators}). \textbf{Panel (c):} \emph{different agents generate
different operators} --- a Planner spends nearly half its segments Planning,
a Decider is dominated by Concluding, and mathematical and knowledge roles
never Review or Test at all. \textbf{Panel (b):} \emph{different operators use
different experts} --- the operator-conditional expert distributions differ
on all nine tasks we measured, at $2.2$--$19\times$ a permutation null that
shuffles operator labels within each agent ($p \le .006$), so the effect is
not agent identity in disguise. \textbf{Panel (a):} \emph{different agents use different experts}, the
composition of the two. A prefill control supports the
direction of causation: on identical token text, agents do not diverge ---
the prompt acts by changing what the agent writes, and what it writes
determines where it routes. This is also what makes prediction possible:
because the prompt determines the behavior and the behavior determines the
routing, the prompt carries the information needed to anticipate an agent's
experts before generation begins --- the intuition DEP is built on.

This is the requirement the rest of the paper answers: agents in different
roles and on different tasks call for \emph{specialized} expert masks ---
Section~\ref{sec:heterogeneity} shows that masks of two agents in one
workflow can overlap no more than masks of two unrelated benchmarks --- and
since calibrating one mask per configuration reinstates the very cost
pruning was meant to remove, the specialization must be computed from
the agent's prompt rather than measured from a corpus. That is the method of
Section~\ref{sec:method}. Appendix~\ref{app:related} situates this position
among static compression, input-conditioned sparsity, and MoE serving
systems.

\section{Methods}
\label{sec:method}

\subsection{Problem formulation}
\label{sec:setting}

We consider serving a frozen Mixture-of-Experts (MoE) language model
$\mathcal{M}$ with $L$ layers, each containing $E$ routed experts of which
the router activates the top-$k$ per token. The workload is not a single
task but a collection of multi-agent workflows
$\{\mathcal{W}_1, \dots, \mathcal{W}_M\}$: each workflow
$\mathcal{W} = \{a_1, \dots, a_n\}$ is a program over agents, where
agent $a$ is defined by a system prompt $p_a$ and is invoked one or more times
while the workflow processes an input problem $q$. All agents of all workflows
are served by the backbone $\mathcal{M}$; they differ only in their
prompts and intermediate context. Crucially, the set of workflows is open:
workflows are edited, re-searched, and added over the life of a deployment, so
a serving-time method cannot assume it has seen every configuration it will
serve.

Expert pruning restricts each MoE layer $\ell$ to a subset
$\mathcal{S}_\ell \subseteq \{1,\dots,E\}$ of size $K$, the \emph{retention}
$\rho = K/E$. Static methods choose a single
$\{\mathcal{S}_\ell\}_{\ell=1}^{L}$ offline from calibration data and apply it to every
request of every workflow. DEP replaces the calibrated subset with a
learned masking function
\begin{equation}
m_\theta(p_a, q) \;\in\; \{0,1\}^{L \times E},
\qquad \textstyle\sum_e m_\theta(p_a, q)_{\ell e} = K \ \ \forall \ell,
\label{eq:maskfn}
\end{equation}
computed once before an agent's turn from the text that defines the request
--- a mask \emph{specialized} to that request --- and held constant for the
whole turn.

\subsection{One-shot mask prediction}
\label{sec:predictor}

DEP's \emph{mask predictor} $f_\theta$, orders of magnitude smaller than
the backbone, maps the pair $(p_a, q)$ to per-layer expert scores ---
operationalizing the intuition of Section~\ref{sec:intro} that the prompts
configuring an agent already specify what it will do and what knowledge that
demands, and therefore suffice as the mask's sole input.
Following the sparsity-predictor design of instruction-following pruning
\citep{hou2025ifpruning}, $f_\theta$ consists of a small pretrained text
encoder and a shallow MLP head: the concatenated text $p_a \oplus q$ is
encoded, the final-token hidden state $h \in \mathbb{R}^{d}$ is pooled, and
the head produces scores
\begin{equation}
S \;=\; W_2\, \phi(W_1 h) \;\in\; \mathbb{R}^{L \times E},
\label{eq:scores}
\end{equation}
with $\phi$ a pointwise nonlinearity and the head's $LE$-dimensional output
reshaped to $L \times E$. The mask of Eq.~\ref{eq:maskfn} keeps the top-$K$
experts per layer by thresholding at the $K$-th largest score:
\begin{equation}
m_{\ell e} \;=\; \mathbf{1}\!\left[ S_{\ell e} \ge \tau_K(S_{\ell}) \right],
\qquad \tau_K(S_\ell) = \text{$K$-th largest entry of } S_\ell .
\label{eq:topk}
\end{equation}
Prediction is a single forward pass of the small encoder per (agent, problem),
amortized over the agent's entire generation. Because the predictor reads
prompt text rather than an identifier drawn from a fixed roster,
nothing ties it to the agents it was trained on: any new prompt yields a
mask, in one shot.

\paragraph{Applying the mask.}
The mask is installed on the frozen backbone at the router: logits of masked
experts are set to $-\infty$ before the router's softmax-and-top-$k$, so
routing renormalizes over the $K$ kept experts and never selects a pruned one.
Per-token top-$k$ routing is otherwise unchanged, model weights are untouched,
and we require $K \ge k$ so that routing remains well defined. This
routing-level view is what accuracy evaluations use; the memory mechanism that
realizes the saving is the serving runtime of Section~\ref{sec:serving}.

\subsection{Predictor training}
\label{sec:training}

The predictor is trained on \emph{workflow transcripts}: each training
workflow is executed by the dense backbone on a calibration split disjoint
from evaluation, and every agent call contributes a triple (system prompt
$p_a$, input $q$, response $r$) attributed to the node that produced it. The
resulting corpus pools all training workflows, so one predictor learns from
the union rather than one artifact per configuration. The backbone stays
frozen; only $\theta$ is trained. Two complementary signals
supervise the scores: one says which experts \emph{mattered}, the other
measures what a candidate mask \emph{breaks}.

\paragraph{Impact-coverage target.}
For each transcript we compute a per-example expert-saliency tensor with the
REAP criterion \citep{lasby2025reap}: for layer $\ell$ and expert $e$,
\begin{equation}
\sigma_{\ell e} \;=\; \frac{1}{|\mathcal{T}_{\ell e}|}
\sum_{t \in \mathcal{T}_{\ell e}} g_{\ell e}(x_t)\,
\bigl\lVert \mathrm{FFN}_{\ell e}(x_t) \bigr\rVert_2 ,
\label{eq:saliency}
\end{equation}
where $\mathcal{T}_{\ell e}$ is the set of response tokens routed to $e$ and
$g_{\ell e}(x_t)$ the router gate weight --- i.e.\ how much the expert actually
\emph{contributes} to this example's output, not merely how often it is
routed. Binarizing $\sigma$ per layer at top-$K$ gives a per-example target
mask $m^\star$, and the predictor is trained with binary cross-entropy,
$\mathcal{L}_{\mathrm{impact}} = \mathrm{BCE}(\mathrm{sigmoid}(S), m^\star)$. This target is
cheap and per-example, but it scores experts in isolation and never sees the
joint effect of removing $E{-}K$ of them at once.

\paragraph{Self-distillation.}
The second signal closes that gap by measuring masking damage directly: with
teacher the unmasked backbone and student the same backbone under the
predicted mask,
\begin{equation}
\mathcal{L}_{\mathrm{KD}} \;=\; \frac{1}{|\mathcal{R}|} \sum_{t \in \mathcal{R}}
\mathrm{KL}\!\left( P_{\mathcal{M}}(\cdot \mid x_{<t}) \;\big\Vert\;
P_{\mathcal{M} \odot \tilde m}(\cdot \mid x_{<t}) \right),
\label{eq:kd}
\end{equation}
averaged over the response tokens $\mathcal{R}$ of the transcript. The teacher is
the model itself, so no labels are needed and the loss directly optimizes the
quantity deployment cares about --- how little the masked model's predictive
distribution moves. To make Eq.~\ref{eq:kd} differentiable in $\theta$, the
hard top-$K$ of Eq.~\ref{eq:topk} is relaxed to a per-layer soft mask
$\tilde m_{\ell e} = \mathrm{sigmoid}\!\left( (S_{\ell e} - \tau_K(S_\ell)) / T \right)$
whose temperature $T$ is annealed toward the hard mask over training; at
inference we use the hard top-$K$ of the same scores. The total objective
mixes the two signals,
\begin{equation}
\mathcal{L} \;=\; \alpha\, \mathcal{L}_{\mathrm{KD}}
\;+\; (1-\alpha)\, \mathcal{L}_{\mathrm{impact}} ,
\label{eq:loss}
\end{equation}
with mixing weight $\alpha$. One predictor is trained per retention level on
the pooled corpus; the same checkpoint then serves every agent of every
workflow at that retention, seen or unseen.

\subsection{Dynamic inference and delta loading}
\label{sec:serving}

\paragraph{Inference.}
A request arrives as (agent prompt, input). The serving engine first obtains
the request's mask: one predictor forward if it is new, or a cache hit ---
masks are content-addressed by $(p_a, q)$, so repeated calls of an agent on
the same problem, and batches grouped by agent, reuse one mask. The backbone
then decodes the entire turn under that mask: per-token top-$k$ routing runs
natively, restricted to the $K$ kept experts. Nothing about the mask changes
mid-turn, so the inner decoding loop is identical to serving a statically
pruned model. Resident memory is the non-expert weights plus a $\rho$
fraction of the experts --- which dominate an MoE's footprint, so memory
shrinks nearly in proportion to $\rho$ --- and adapting to a new workflow,
agent, or edited prompt costs one small predictor forward rather than the
recalibration pass a static mask requires.

\paragraph{Delta loading.}
The memory saving comes from keeping only the kept experts on the
accelerator. Expert weights live as per-expert tensors in host memory; each
layer holds a resident buffer of $K$ slots plus an index map from global
expert ids to slots, and the router is masked to the resident set so an
absent expert can never be selected. When the mask changes --- typically at
an agent hand-off --- the engine copies only the set difference between the
required and resident experts into the slots that evicted ones vacate, and
updates the map; experts shared by consecutive masks never move. That delta
stays small because the masks of agents in one workflow overlap substantially
(Section~\ref{sec:heterogeneity}), and it is largely hideable because the
workflow's topology is known in advance: the next agent's mask can be
predicted and its delta prefetched while the current turn is still decoding,
and batching by agent amortizes whatever remains.

\section{Experiments}
\label{sec:experiments}

\subsection{Experimental setup}
\label{sec:setup}

\paragraph{Models and workflows.}
Our main backbone is Qwen3-30B-A3B \citep{yang2025qwen3} (48 MoE layers,
$E{=}128$ routed experts, top-$8$ routing), evaluated at retention
$\rho{=}50\%$ ($K{=}64$) and the aggressive $\rho{=}25\%$ ($K{=}32$); the
predictor encoder is Qwen2-0.5B. For agentic coding (Section~\ref{sec:swebench})
we use Qwen3-Coder-30B-A3B-Instruct and, to test scale,
Qwen3-Coder-480B-A35B-Instruct-FP8 (62 layers, $E{=}160$); to test a
different MoE family we additionally use DeepSeek-V2.5-236B (59 layers,
$E{=}160$ routed, top-$6$, plus 2 shared experts). Workflows are
produced by automated search rather than hand-designed: per task, an AFlow
MCTS search \citep{zhang2025aflow} with the target backbone as executor, from
which we select the best-scoring discovered workflow
(Appendix~\ref{app:setup}).

\paragraph{Tasks and protocol.}
Six tasks span mathematical reasoning, code generation, and knowledge-seeking
QA: MATH-500 \citep{hendrycks2021math,lightman2023verify}, GSM8K
\citep{cobbe2021gsm8k}, HumanEval+ and MBPP+
\citep{chen2021codex,austin2021mbpp,liu2023evalplus}, MMLU \citep{hendrycks2021mmlu},
and HotpotQA \citep{yang2018hotpotqa}, each with one discovered workflow;
$n{=}100$ held-out problems per seed over 3 seeds (HumanEval+ caps at 69).
Evaluation pools are disjoint from search-validation, predictor-training, and
calibration data. Agentic coding uses a fixed 50-instance subset of SWE-bench
Verified \citep{jimenez2024swebench, chowdhury2024swebenchverified} under the
official harness in an oracle-assisted setting, so the object of comparison is
the \emph{relative} ordering of policies, not leaderboard-absolute numbers.

\paragraph{Baselines.}
All policies share the identical frozen backbone and workflow, and every
calibration-based baseline is calibrated on the \emph{same} transcripts that
train our predictor: \textbf{dense} (unpruned reference), \textbf{direct}
(unpruned single call --- the workflow's own contribution), \textbf{freq}
(static mask by routing frequency), \textbf{REAP} \citep{lasby2025reap}
(static mask by the saliency of Eq.~\ref{eq:saliency}, a state-of-the-art
static pruning criterion), \textbf{HC-SMoE} \citep{chen2024hcsmoe} (expert
\emph{merging} by agglomerative clustering), and \textbf{DEP} (ours; the
per-request predictor mask of Section~\ref{sec:method}). One predictor per
retention is trained on the pooled six-workflow transcript corpus
($\sim$1.5k examples balanced across tasks) with the objective of
Eq.~\ref{eq:loss}; accuracy runs mask routing only, while memory and latency
are measured separately with the delta-loading runtime of
Section~\ref{sec:serving}. All remaining details --- workflow structures,
contamination checks, training hyperparameters, efficiency protocol, and
hardware --- are in Appendix~\ref{app:setup}; ablations of the training
recipe are in Appendix~\ref{app:ablations}.

\subsection{Dynamic versus static expert pruning}
\label{sec:main-results}

Table~\ref{tab:main} compares all policies on six benchmarks, each served by a
workflow discovered for it by AFlow, at both retention levels.

\begin{table}[t]
\centering
\caption{\textbf{DEP gives the best overall accuracy at both retentions, and
is the only policy still usable at $\rho{=}25\%$.} AFlow-discovered
multi-agent workflows (Qwen3-30B-A3B); DEP against static masks calibrated on
the \emph{identical} union. Values are \% (pass@1; HotpotQA SQuAD-F1; MMLU
accuracy), mean\,$\pm$\,std; protocol in Section~\ref{sec:setup}.}
\label{tab:main}
\tabfont
\setlength{\tabcolsep}{5.2pt}
\begin{tabular}{l cccccc | c}
\toprule
\textbf{Method} & MATH-500 & GSM8K & HumanEval+ & MBPP+ & MMLU & HotpotQA & \textit{Mean} \\
\midrule
Dense (56.9\,GB)
 & 68.3\,\tiny$\pm$8.6 & 97.3\,\tiny$\pm$1.2 & 92.8\,\tiny$\pm$1.4
 & 80.0\,\tiny$\pm$6.9 & 82.7\,\tiny$\pm$1.5 & 71.9\,\tiny$\pm$1.7
 & 82.2\,\tiny$\pm$1.0 \\
\midrule
\multicolumn{8}{l}{\emph{50\% retention (keep 64/128; 29.9\,GB resident)}} \\
Freq.
 & 62.3\,\tiny$\pm$9.0 & \textbf{95.3}\,\tiny$\pm$1.5 & 2.9\,\tiny$\pm$1.4
 & 3.0\,\tiny$\pm$1.0 & 51.0\,\tiny$\pm$2.6 & \textbf{65.4}\,\tiny$\pm$3.7
 & 46.7\,\tiny$\pm$1.9 \\
REAP
 & 40.7\,\tiny$\pm$6.8 & 71.3\,\tiny$\pm$4.5 & 78.7\,\tiny$\pm$0.8
 & 45.7\,\tiny$\pm$4.0 & 40.3\,\tiny$\pm$4.0 & 47.3\,\tiny$\pm$3.0
 & 54.0\,\tiny$\pm$1.6 \\
HC-SMoE
 & 60.3\,\tiny$\pm$6.4 & 84.0\,\tiny$\pm$3.0 & 75.8\,\tiny$\pm$2.2
 & 75.0\,\tiny$\pm$7.8 & 48.3\,\tiny$\pm$0.6 & 60.4\,\tiny$\pm$1.6
 & 67.3\,\tiny$\pm$1.5 \\
\textbf{DEP}
 & \textbf{67.3}\,\tiny$\pm$10.6 & \textbf{95.3}\,\tiny$\pm$2.3 & \textbf{84.5}\,\tiny$\pm$0.8
 & \textbf{76.7}\,\tiny$\pm$2.5 & \textbf{70.7}\,\tiny$\pm$4.7 & 60.7\,\tiny$\pm$3.2
 & \textbf{75.9}\,\tiny$\pm$2.3 \\
\midrule
\multicolumn{8}{l}{\emph{25\% retention (keep 32/128; 16.4\,GB resident)}} \\
Freq.
 & 4.0\,\tiny$\pm$1.0 & 2.3\,\tiny$\pm$0.6 & 0.0\,\tiny$\pm$0.0
 & 0.0\,\tiny$\pm$0.0 & 0.7\,\tiny$\pm$1.2 & 0.2\,\tiny$\pm$0.0
 & 1.2\,\tiny$\pm$0.2 \\
REAP
 & 5.3\,\tiny$\pm$1.5 & 7.7\,\tiny$\pm$4.0 & 0.0\,\tiny$\pm$0.0
 & 12.3\,\tiny$\pm$2.1 & 10.7\,\tiny$\pm$2.5 & 4.3\,\tiny$\pm$0.9
 & 6.7\,\tiny$\pm$1.4 \\
HC-SMoE
 & 2.7\,\tiny$\pm$1.5 & 0.7\,\tiny$\pm$1.2 & 0.0\,\tiny$\pm$0.0
 & 0.3\,\tiny$\pm$0.6 & 26.3\,\tiny$\pm$1.5 & 5.0\,\tiny$\pm$0.6
 & 5.8\,\tiny$\pm$0.1 \\
\textbf{DEP}
 & \textbf{51.3}\,\tiny$\pm$7.2 & \textbf{87.7}\,\tiny$\pm$4.7 & \textbf{78.3}\,\tiny$\pm$2.9
 & \textbf{61.7}\,\tiny$\pm$6.5 & \textbf{48.0}\,\tiny$\pm$1.0 & \textbf{53.7}\,\tiny$\pm$2.7
 & \textbf{63.4}\,\tiny$\pm$1.6 \\
\bottomrule
\end{tabular}
\end{table}

DEP obtains the best overall accuracy of any pruning policy at both
retention levels, and leads on most individual benchmarks. The more telling
pattern, though, is not where the margins are but where the static baselines
break down: each of them collapses on some subset of the workload while
remaining respectable on the rest. We believe this is what a single mask
costs. A static criterion must score every expert on a mixture of tasks and
agents at once, so experts that matter to only part of that mixture look
unimportant on average and are dropped --- and whichever task depended on
them fails outright rather than degrading. DEP never has to make that
compromise: it emits a separate mask per agent and task, so an expert only
has to matter to the request being served in order to be kept. That it can
do so from the system and task prompts alone is the part we would emphasize
--- the text that configures an agent is enough to identify the experts that
agent will need, before it has generated a token.

The same effect sharpens at $\rho{=}25\%$, where DEP's margin over every
baseline is at its largest. Halving the budget again leaves no room for a
mask to hedge: a subset chosen to be adequate everywhere is now too small to
be sufficient anywhere, and the static policies lose the experts their tasks
actually depend on, which shows up as collapsed rather than merely weakened
generation (Section~\ref{sec:efficiency}). Spending the same budget on one
agent's experts instead keeps what that agent depends on, and the kept set
stays coherent even when it is small. DEP does give up real accuracy against
dense here --- this is graceful degradation, not free compression --- but the
regime is informative in its own right: it suggests that a multi-agent
decomposition is itself a form of specialization that can be exploited for
efficiency, since an agent responsible for one narrow part of a task needs
correspondingly fewer experts, and can therefore be served far more sparsely
than a model expected to do everything.

\subsection{Generalization to unseen workflows and benchmarks}
\label{sec:generalization}

Section~\ref{sec:intro} claimed that DEP masks new configurations in one
shot, because nothing ties a prompt-conditioned predictor to the workflows it
was trained on. We test that claim directly with a single universal
predictor. Unlike the other experiments, which serve one discovered workflow
per task, here the predictor is trained on transcripts from a \emph{pool} of
32 discovered workflows spanning three tasks (full rosters in
Appendix~\ref{app:rosters}), then evaluated \emph{without any retraining} on
a ladder of increasing distance from training: \textbf{unseen
structures} of workflow families seen in training; an entire \textbf{unseen
workflow family}; \textbf{unseen benchmarks} served by unseen workflows
(SVAMP \citep{patel2021svamp}, near the training math distribution; MBPP,
far); and an
\textbf{unseen domain} (MMLU).
Every static baseline is calibrated once on the same training transcripts and
reused unchanged at every rung --- exactly the situation a static mask faces
when the workload changes after calibration.

\begin{table}[t]
\centering
\caption{\textbf{One predictor, fixed after training, keeps working as the
workload moves away from it.} Generalization ladder: one universal predictor
and one static calibration, both fixed after training on a three-task
workflow pool, evaluated at increasing distance from training. Accuracy \% (HotpotQA F1),
30 problems per cell, mean\,$\pm$\,std over 3 seeds. \textbf{Bold} = best,
\underline{underline} = second-best pruned method per row and retention.}
\label{tab:gen}
\tabfont
\setlength{\tabcolsep}{1.0pt}
\begin{tabular}{ll c cccc cccc}
\toprule
& & & \multicolumn{4}{c}{$\rho = 50\%$} & \multicolumn{4}{c}{$\rho = 25\%$} \\
\cmidrule(lr){4-7} \cmidrule(lr){8-11}
 & task & Dense & Freq. & REAP & HC-SMoE & DEP & Freq. & REAP & HC-SMoE & DEP \\
\midrule
\multirow{3}{*}{\shortstack[l]{Unseen\\structure}}
 & HotpotQA   & 68.0\tiny$\pm$6.6 & 12.9\tiny$\pm$1.5 & 12.3\tiny$\pm$2.7 & \underline{54.5\tiny$\pm$5.5} & \textbf{65.0\tiny$\pm$2.8} & 0.4\tiny$\pm$0.4 & \underline{5.3\tiny$\pm$3.3} & 0.1\tiny$\pm$0.1 & \textbf{38.3\tiny$\pm$0.6} \\
 & HumanEval+ & 94.3\tiny$\pm$3.5 & 39.0\tiny$\pm$2.6 & 59.3\tiny$\pm$1.5 & \underline{82.3\tiny$\pm$1.5} & \textbf{83.4\tiny$\pm$4.1} & 0.0\tiny$\pm$0.0 & \underline{7.7\tiny$\pm$2.5} & 0.0\tiny$\pm$0.0 & \textbf{9.7\tiny$\pm$3.2} \\
 & MATH       & 68.3\tiny$\pm$3.3 & 0.0\tiny$\pm$0.0 & 2.8\tiny$\pm$2.5 & \underline{45.8\tiny$\pm$8.2} & \textbf{55.0\tiny$\pm$2.4} & 0.0\tiny$\pm$0.0 & 0.0\tiny$\pm$0.0 & 0.0\tiny$\pm$0.0 & \textbf{9.4\tiny$\pm$1.0} \\
\midrule
\multirow{3}{*}{\shortstack[l]{Unseen\\family}}
 & HotpotQA   & 67.5\tiny$\pm$4.3 & 9.9\tiny$\pm$0.1 & 29.5\tiny$\pm$8.9 & \textbf{56.4\tiny$\pm$7.2} & \underline{53.1\tiny$\pm$3.2} & 0.0\tiny$\pm$0.0 & \underline{3.1\tiny$\pm$1.2} & 0.0\tiny$\pm$0.0 & \textbf{44.4\tiny$\pm$2.6} \\
 & HumanEval+ & 92.0\tiny$\pm$1.9 & 58.7\tiny$\pm$1.7 & 74.0\tiny$\pm$2.1 & \underline{87.6\tiny$\pm$3.4} & \textbf{88.2\tiny$\pm$3.9} & 0.0\tiny$\pm$0.0 & \underline{27.5\tiny$\pm$3.2} & 0.0\tiny$\pm$0.0 & \textbf{28.0\tiny$\pm$2.3} \\
 & MATH       & 68.3\tiny$\pm$0.8 & 30.0\tiny$\pm$1.3 & 2.2\tiny$\pm$1.4 & \underline{48.9\tiny$\pm$2.5} & \textbf{54.4\tiny$\pm$1.4} & 0.0\tiny$\pm$0.0 & \underline{1.1\tiny$\pm$0.6} & 0.0\tiny$\pm$0.0 & \textbf{21.7\tiny$\pm$3.9} \\
\midrule
\multirow{2}{*}{\shortstack[l]{Unseen\\benchmark}}
 & SVAMP      & 91.9\tiny$\pm$3.6 & 28.3\tiny$\pm$3.9 & 19.4\tiny$\pm$5.5 & \textbf{87.8\tiny$\pm$3.1} & \underline{82.8\tiny$\pm$0.8} & 1.1\tiny$\pm$0.6 & \underline{12.8\tiny$\pm$2.4} & 1.1\tiny$\pm$0.6 & \textbf{51.7\tiny$\pm$5.5} \\
 & MBPP       & 63.3\tiny$\pm$5.2 & 24.4\tiny$\pm$3.4 & \underline{47.8\tiny$\pm$11.0} & \textbf{51.9\tiny$\pm$3.6} & 46.7\tiny$\pm$1.6 & 0.0\tiny$\pm$0.0 & \textbf{27.2\tiny$\pm$5.5} & 0.6\tiny$\pm$0.8 & \underline{8.3\tiny$\pm$2.4} \\
\midrule
\shortstack[l]{Unseen\\domain}
 & MMLU       & 68.3\tiny$\pm$2.4 & 12.2\tiny$\pm$1.6 & 13.3\tiny$\pm$2.5 & \underline{33.3\tiny$\pm$3.6} & \textbf{48.9\tiny$\pm$3.1} & 0.0\tiny$\pm$0.0 & \underline{10.0\tiny$\pm$1.7} & 1.7\tiny$\pm$0.8 & \textbf{15.6\tiny$\pm$1.6} \\
\bottomrule
\end{tabular}
\end{table}

The predictor transfers. With no retraining and no recalibration it is the
strongest policy on most rungs at both retentions (Table~\ref{tab:gen}), and at $\rho{=}50\%$ it
comes close to the dense reference on several of them: a mask produced for a
workflow the predictor has never seen is still a mask worth serving. This is
the claim of Section~\ref{sec:intro} tested at its weakest point --- the
prompts here describe agents, structures, and in two rungs whole benchmarks
and domains that were absent from training --- and the prompts still carry
enough to pick a usable expert set.

Two things temper the result. Expert merging remains competitive while the
budget is loose, leading on a minority of rungs at $\rho{=}50\%$ and on the
unseen benchmarks in particular; averaging experts rather than committing to
a subset is a reasonable hedge when the workload is unfamiliar and the budget
can absorb it. And with 30 problems per cell the seed spread is not
negligible: on a few rungs the gap between the best and second-best policy is
smaller than a seed standard deviation, so the ordering across the ladder,
rather than any individual cell, is what this table supports. At
$\rho{=}25\%$ the picture sharpens: merging collapses, and DEP leads on every
non-code rung by margins well outside the seed spread. Code is the
exception, REAP stays competitive there, and on the unseen code benchmark
it is the stronger policy. We read that contrast as a
difference in what the two kinds of artifact can express: a calibrated mask
answers ``which experts does \emph{this} workload need'', a question with no
answer once the workload changes, whereas the predictor has learned ``which
experts does an agent \emph{like this} need'', which still has an answer
for a configuration it has never seen.

\subsection{Analysis: predicted masks specialize by task and role}
\label{sec:heterogeneity}

\begin{figure}[t]
\centering
\includegraphics[width=0.338\linewidth]{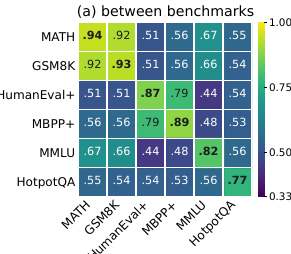}\hfill
\includegraphics[width=0.303\linewidth]{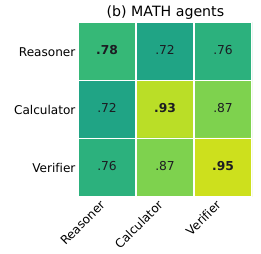}\hfill
\includegraphics[width=0.314\linewidth]{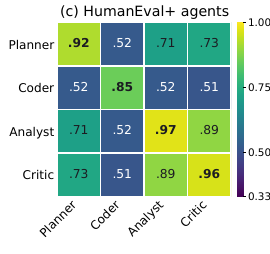}
\caption{\textbf{Predicted masks organize themselves by task and by role.}
Jaccard overlap of predicted kept-expert masks at $\rho{=}50\%$
(chance $=$ 0.33). \textbf{(a)} Between benchmarks; \textbf{(b, c)} between
the agents of the MATH and HumanEval+ teams. Diagonals: the same row on
different problems. Measured on hand-crafted agent teams, not the discovered
workflows of Table~\ref{tab:main}.}
\label{fig:heterogeneity}
\end{figure}

How does the predictor actually spend its budget?
Figure~\ref{fig:heterogeneity} looks inside the predicted masks, measuring
the overlap of the kept-expert sets between benchmarks and between the agents
of one team. The setting is hand-crafted agent teams that follow the role
decompositions in widespread use in multi-agent frameworks
\citep{wu2023autogen, hong2024metagpt}. The masks organize themselves by
expertise. Between benchmarks, domains that share skills share experts:
MATH and GSM8K keep largely the same set, as do HumanEval+ and MBPP+, while
overlap across domain boundaries drops sharply. Between agents, the Coder
keeps a visibly distinct expert set while the Analyst and Critic share most
of theirs --- mirroring the routed-usage picture of
Figure~\ref{fig:motivation}(a), where the Coder was likewise the outlier
role. None of this structure is supervised: the predictor is trained only
with the objective of Eq.~\ref{eq:loss}, yet it allocates experts that track
what each task and role actually requires --- the specialization that
Section~\ref{sec:motivation} argued multi-agent serving demands.

This heterogeneity is consistent with the pattern in Table~\ref{tab:main},
though it is measured on these teams rather than on the discovered workflows
served there. Where even closely related requests keep
overlapping-but-different sets, any single static subset misfits a large
fraction of the request stream --- harmlessly while the budget has slack,
fatally at $\rho{=}25\%$: a static mask is in effect an average of
incompatible masks, and the sparser the budget, the more that averaging
would cost.

\subsection{Efficiency: memory, wall-time, and the failure mode of wrong masks}
\label{sec:efficiency}

\begin{table}[t]
\centering
\caption{\textbf{A wrong mask wastes more time rambling than a right mask
spends swapping experts.} Efficiency at $\rho{=}50\%$ over three workflows
(HotpotQA, HumanEval+, MATH; 3 random 50-problem seeds), not comparable with
Table~\ref{tab:main}; per task in Table~\ref{tab:efficiency-full}. Tokens
and times are measured; $\Delta$-load is an analytic \emph{upper bound} on
the per-agent-step swap cost, which static masks never pay.}
\label{tab:efficiency}
\tabfont
\setlength{\tabcolsep}{9.2pt}
\begin{tabular}{l cccccc}
\toprule
policy & memory (GB) & accuracy & tokens/prob & generation (s) & $\Delta$-load (s) & total (s) \\
\midrule
dense & 56.9 & \textbf{0.78} & 1905 & 3239 & --- & 3239 \\
\textbf{DEP} & \textbf{29.9} & 0.76 & \textbf{1872} & 3275 & +42 & \textbf{3317} \\
REAP  & \textbf{29.9} & 0.53 & 2077 & 3803 & 0 & 3803 \\
freq  & \textbf{29.9} & 0.27 & 3365 & 5457 & 0 & 5457 \\
\bottomrule
\end{tabular}
\end{table}

With delta loading (Section~\ref{sec:serving}), resident memory shrinks
nearly in proportion to the retention, and Figure~\ref{fig:single-vs-multi}
(right) traces the resulting accuracy--memory frontier: DEP holds accuracy at
footprints where both static masks have already collapsed, reconverging with
them only at mild compression, where pruning is nearly free for everyone.
DEP's overheads --- one predictor forward per request and an expert swap per
agent hand-off --- add about one percent to the measured generation wall in
aggregate, and under ten percent on any single task, when charged at their
analytic worst case (Table~\ref{tab:efficiency}). Even with that upper bound
applied to DEP and no swap cost at all charged to the static masks, DEP
still finishes \emph{faster} overall than they do. The reason is token waste: a mask that has lost
essential experts does not answer more briefly, it degenerates, emitting
repetitive tokens until the generation cap (Appendix~\ref{app:collapse}) ---
a wrong mask loses more time rambling than a right mask spends loading, so
the accuracy, token, and latency gaps in Table~\ref{tab:efficiency} all share
one cause.

\subsection{Ablation analysis: scale and architecture agnosticism}
\label{sec:swebench}

Finally we ask whether the result depends on the setting that produced it,
along three axes: a multi-agent benchmark outside the training suite, a much
larger backbone, and a different MoE architecture. On SWE-bench Verified
(Table~\ref{tab:swe}) DEP is the strongest pruning policy at both model
scales, on the 30B coder and on the 16$\times$ larger 480B one, whose dense
entry is a floor because four instances hit harness image errors. The setting
is oracle-assisted, so the ordering rather than the absolute rate is the
result --- and it is the same ordering Table~\ref{tab:main} produced on a
different task family. Even the failure mode recurs: the frequency mask does
not go silent but emits degenerate patches, garbage diffs in place of garbage
tokens.

\begin{table}[t]
\begin{minipage}[t]{0.49\linewidth}
\centering
\caption{\textbf{The ordering carries over to agentic coding at both model
scales.} SWE-bench Verified resolve rate (\%) at $\rho{=}50\%$.}
\label{tab:swe}
\tabfont
\setlength{\tabcolsep}{7.8pt}
\begin{tabular}{l cc}
\toprule
\textbf{Method} & Coder-30B & Coder-480B \\
\midrule
Dense (workflow)     & 40.0 & 56.0 \\
Direct (single call) & 12.0 & 50.0 \\
\midrule
Freq.         & 2.0  & 0.0 \\
REAP          & 22.0 & 44.0 \\
\textbf{DEP} & \textbf{30.0} & \textbf{50.0} \\
\bottomrule
\end{tabular}
\end{minipage}\hfill
\begin{minipage}[t]{0.49\linewidth}
\centering
\caption{\textbf{It also holds on a different MoE family.}
DeepSeek-V2.5-236B, mean over five tasks (\%); per task in
Table~\ref{tab:arch-full}.}
\label{tab:arch}
\tabfont
\setlength{\tabcolsep}{7.8pt}
\begin{tabular}{l cc}
\toprule
\textbf{Method} & $\rho{=}50\%$ & $\rho{=}25\%$ \\
\midrule
Dense (workflow)     & \multicolumn{2}{c}{70.7} \\
Direct (single call) & \multicolumn{2}{c}{71.4} \\
\midrule
Freq.         & 33.0 & 0.8 \\
REAP          & 46.5 & 1.5 \\
\textbf{DEP} & \textbf{51.7} & \textbf{6.9} \\
\bottomrule
\end{tabular}
\end{minipage}
\end{table}

Nothing in DEP's design is specific to one MoE family: the mask acts at the
router, and the serving runtime assumes only per-expert weight tensors.
Table~\ref{tab:arch} tests that on DeepSeek-V2.5-236B, a different MoE
lineage with its own routing width and a pair of shared, always-active
experts we leave unmasked. DEP again gives the best overall accuracy of any
pruning policy at both retentions, and per task
(Table~\ref{tab:arch-full}) the frequency mask again breaks first on code.
What does not transfer is the retention level: pruning costs more of the
dense accuracy here than on Qwen3, and at $\rho{=}25\%$ every policy becomes
unusable, DEP included, even though it keeps the largest margin. How
aggressively a backbone tolerates pruning is thus a property of that
backbone, to be measured; what carries over is which policy to use once the
budget is fixed.

\section{Conclusion}
\label{sec:conclusion}

Serving many agents from one backbone breaks the assumption expert pruning
rests on: agents in different roles and on different tasks recruit measurably
different experts, so a mask averaged over all of them is fitted to none, and
pruning has to become dynamic rather than better calibrated. DEP makes it
dynamic by prediction rather than by calibration: a small predictor, trained
once on workflow transcripts and frozen thereafter, emits a per-request mask
in a single forward pass. The observation we would most want carried forward
is that the prompts already suffice --- the system and task prompts that
configure an agent are by themselves enough to identify the experts it will
need, before it has generated a token and even when neither the agent nor its
workflow appeared in training. Our experiments bear that out across
workloads, model scales and MoE architectures, and our analyses trace the
behavior to its cause: masks that specialize by task and by role, and a
wrong mask that costs more in degenerate generation than a right one costs in
expert swaps. The price is a single predictor training run, and how
aggressively a given backbone tolerates pruning must still be measured rather
than assumed (Appendix~\ref{app:limitations}). Role specialization, then,
cuts both ways: the division of labor that multi-agent systems adopt for
capability also narrows what each agent must know, and a narrower need is one
that far fewer experts can serve.


\section*{AI use statement}

Generative AI was used as a writing and coding assistant: to help draft and
edit prose. The research idea, the method, and the claims are solely the
authors' own. All AI-assisted work was reviewed by the authors --- every
reported number was produced by our own runs and checked against the stored
prediction files, and all code was read and tested before use. We take
responsibility for the final content of this work.

\section*{Reproducibility Statement}

Section~\ref{sec:method} specifies the masking function, training objective
and serving procedure in full (Eqs.~\ref{eq:maskfn}--\ref{eq:loss}), and
Appendix~\ref{app:setup} gives workflow discovery, the predictor's training
configuration and compute, the efficiency protocol and the hardware;
Appendices~\ref{app:rosters} and \ref{app:ablations} give the generalization
rosters and the recipe's sensitivity. Two conditions matter most for
comparability and are stated in Section~\ref{sec:setup}: every
calibration-based baseline is calibrated on the same transcripts that train
our predictor, and evaluation pools are disjoint from search-validation,
predictor-training and all calibration data, verified by text-overlap checks.
Code for predictor training, the masking and delta-loading runtime, and the
evaluation harness is provided as supplementary material.

\bibliography{iclr2027_conference}
\bibliographystyle{iclr2027_conference}

\appendix
\section{Related work}
\label{app:related}

\paragraph{Static MoE expert compression.}
Static expert pruning scores experts on a calibration corpus and
freezes the result: by routing frequency, by the reconstruction error of
removing an expert \citep{lu2024naee}, or by router-weighted activation
magnitude \citep{lasby2025reap}, a state-of-the-art static criterion;
HC-SMoE \citep{chen2024hcsmoe} instead \emph{merges} redundant experts by
agglomerative clustering, trading peak accuracy for graceful degradation.
All of these produce a single artifact fitted to the calibration
distribution --- the assumption that Section~\ref{sec:motivation} shows
failing in multi-agent serving --- and all serve as baselines in
Section~\ref{sec:experiments}.

\paragraph{Input-conditioned sparsity.}
Closer to DEP are methods that condition sparsity on the input rather than
on a corpus. Deja Vu \citep{liu2023dejavu} exploits per-token contextual
sparsity in \emph{dense} models, predicting on the fly which attention heads
and MLP neurons a token needs; the prediction is per token and per layer,
so the resident weight set changes continuously. Instruction-following
pruning \citep{hou2025ifpruning}, whose predictor design we adopt
(Section~\ref{sec:predictor}), conditions a dense-model FFN mask on the
instruction, but selects among sparsity patterns for a single-call setting
and fine-tunes the backbone jointly with the predictor. DEP differs in
granularity and setting: it masks \emph{routed experts} of a frozen MoE
backbone, holds the mask fixed for a whole agent turn so that the resident
set is bounded and cacheable, is trained once on pooled multi-agent
transcripts with a self-distillation objective (Eq.~\ref{eq:kd}), and is
evaluated on its central claim --- one-shot masks for agents, workflows,
and tasks never seen in training (Section~\ref{sec:generalization}) ---
which per-call sparsity selection does not address.

\paragraph{MoE offloading and serving systems.}
A complementary line keeps all experts but moves them: quantized expert
offloading with LRU caching and speculative prefetch
\citep{eliseev2023offloading}, activation-aware request-level tracing for
prefetching and caching \citep{xue2024moeinfinity}, and pre-gating that
computes the next layer's expert set one layer ahead
\citep{hwang2024pregated}. These systems chase per-token routing decisions
reactively; DEP inverts the problem --- the whole turn's expert set is known
\emph{before generation begins}, so expert movement becomes a deterministic
per-turn delta that can be prefetched at agent hand-offs
(Section~\ref{sec:serving}) rather than a per-token cache-miss stream.
Weight quantization \citep{frantar2023gptq, lin2024awq} reduces the
\emph{bytes per expert} rather than the number of resident experts; it is
orthogonal to DEP and composable with it, as our FP8 480B experiments
already illustrate.

\paragraph{Multi-agent systems and their efficiency.}
Multi-agent frameworks assign the same backbone differentiated roles
through prompting \citep{wu2023autogen, hong2024metagpt}, and automated
search extends this by discovering workflow programs
\citep{zhang2025aflow}; whether such compositions beat a single
well-prompted model is actively debated \citep{cemri2025mast, kim2025scaling,
jwalapuram2026illusion}, but either way the resulting traffic is
heterogeneous, which is all DEP requires. Systems work on serving LLM
\emph{applications} optimizes scheduling and cross-request state --- e.g.\
exposing inter-request dependencies to the engine \citep{lin2024parrot} ---
while leaving the model weights untouched; DEP addresses the orthogonal
memory axis, deciding \emph{which experts} must be resident for each
agent's turn, and could sit inside such engines.

\section{Experimental details}
\label{app:setup}

\paragraph{Workflow discovery and structure.}
Per task we run the AFlow MCTS search \citep{zhang2025aflow} with a
Claude-Sonnet optimizer and the target backbone as executor, for up to 10
rounds, and select the best-scoring discovered workflow on the search
validation split. The discovered workflows are genuinely heterogeneous
multi-agent programs; e.g.\ the MATH workflow runs five parallel solvers,
self-consistency ensembling, a programmatic cross-check that writes and
executes code, and a verifier (8 model calls per problem). For SWE-bench we
use a fixed four-role workflow (Localizer $\rightarrow$ Programmer
$\rightarrow$ Reviewer $\rightarrow$ Debugger) in an oracle-assisted setting:
gold affected-file paths and failing-test names are provided in the prompt,
which is why absolute resolve rates are not comparable to end-to-end agent
leaderboards.

\paragraph{Discourse-operator taxonomy.}
\label{app:operators}
The operator taxonomy of Section~\ref{sec:motivation} adapts the
segmentation \emph{methodology} of \citet{reasonops2026} but not its
dictionary: that dictionary is built around reflective markers, and our
traces are non-thinking (``wait'' appears in 0.9\% of messages, ``hmm'' in
none), so forced matching produced labels the text does not support. We
instead rediscover the taxonomy from the corpus itself: sentence-opening
pivots are mined from 8{,}125 messages, the 311 accepted pivots are embedded
and clustered, and a silhouette sweep over the number of clusters selects
the nine committal operators used in the paper. The dictionary is then
frozen and applied unchanged to every pool; Figure~\ref{fig:motivation}(c)
reports every role with at least 50 segments. The operator analysis runs on
nine task pools, a superset of the six benchmarks evaluated in
Section~\ref{sec:experiments}.

\paragraph{Contamination control.}
Evaluation pools are disjoint from (i) the AFlow search-validation problems,
(ii) every problem used to collect predictor-training transcripts, and
(iii) every problem used to calibrate any static baseline; disjointness is
verified by text-overlap checks before first evaluation on each task.

\paragraph{Predictor training.}
Transcripts are collected by running each discovered workflow with the dense
backbone on its search-validation split. The six-task corpus is balanced on
usable examples rather than problems (workflow depth differs per task):
379 problems yielding 1{,}542 per-agent-call examples, $\sim$253--261 per
task, over 10 workflow nodes. One predictor per retention is trained with the
objective of Eq.~\ref{eq:loss} ($\alpha{=}0.5$) for 8 epochs: AdamW, learning
rate $10^{-4}$, weight decay 0.01, gradient clipping 1.0, accumulation 4,
head bottleneck 1024 with GELU, soft-top-$K$ temperature annealed
$0.7 \rightarrow 0.15$. Training takes $\sim$3 GPU-hours per retention level
on two 48\,GB GPUs; no component of the backbone is updated.

\paragraph{Efficiency measurement.}
Accuracy runs apply masks at the router (all experts resident), so memory and
latency are measured separately. Resident-memory savings use the real
delta-loading runtime of Section~\ref{sec:serving} --- per-expert weights in
pinned host memory, a resident buffer of $K$ slots per layer with a
global-to-slot index map, and the router masked to the resident set ---
measuring actual peak allocator usage and host-to-device swap latency as a
function of the number of experts changed. Wall-clock generation time and
token counts come from sequential, contention-free runs over three random
50-problem subsets (mean $\pm$ std).

\paragraph{Hardware.}
30B-model experiments ran on 2$\times$48\,GB GPUs (generation and training)
and the 480B model on 3$\times$B200; SWE-bench patches are scored with the
official Docker harness on a separate host.

\section{Ablations: the training recipe is robust and un-tuned}
\label{app:ablations}

All ablations use a three-task subset of the main setting (MATH, HumanEval+,
HotpotQA; discovered workflows, $\rho{=}50\%$, $n{=}50$ held-out problems);
every cell shares identical hyperparameters, and the sweeps were run
\emph{after} the main configuration was fixed --- so the numbers in
Section~\ref{sec:experiments} are not the product of a hyperparameter
search.

\begin{table}[t]
\centering
\caption{Objective ablation ($\rho{=}50\%$, $n{=}50$). Both terms of
Eq.~\ref{eq:loss} are load-bearing.}
\label{tab:ablation-loss}
\tabfont
\begin{tabular}{l cccc}
\toprule
Objective & MATH & HumanEval+ & HotpotQA (F1) & Mean \\
\midrule
Dense reference        & 82.0 & 92.0 & 72.2 & 82.1 \\
\midrule
KD $+$ BCE(REAP) (ours) & 76.0 & \textbf{90.0} & \textbf{71.7} & \textbf{79.2} \\
LM $+$ BCE(REAP)        & 76.0 & 84.0 & 68.4 & 76.1 \\
LM $+$ BCE(norm)        & \textbf{82.0} & 80.0 & 0.0  & 54.0 \\
KD only ($\alpha{=}1$)  & 30.0 & ---  & ---  & ---  \\
\bottomrule
\end{tabular}
\end{table}

\paragraph{Both loss terms are load-bearing.}
Over a $4{\times}5$ grid of distillation objectives $\times$ saliency
criteria, no cell beat the shipped combination
(Table~\ref{tab:ablation-loss}). Removing the criterion term entirely
($\alpha{=}1$, pure self-distillation) collapses MATH from 76.0 to 30.0,
while replacing the KD term with a plain masked language-modeling loss both
degrades the mean and can produce \emph{degenerate} predictors --- the
LM$+$norm cell decodes a single repeated token on 47/50 HotpotQA problems.
Between these extremes the mixing weight is uncritical:
$\alpha \in [0, 0.75]$ varies MATH by $\pm 3$ points with no trend.

\paragraph{The saliency criterion is already near-optimal.}
Among the criteria in the grid, REAP (Eq.~\ref{eq:saliency}) is the best
target; it also approximates an \emph{exact} gradient-based Taylor saliency
better (top-$K$ set agreement 0.854) than routing frequency (0.780), output
norm (0.839), or two purpose-built forward-only Taylor surrogates
(0.796--0.799) --- at zero extra cost, since it is computed in the same
forward pass as the transcripts.

\begin{table}[t]
\centering
\caption{Hyperparameter sweep (screened on MATH, the only task with
headroom; paired on identical problems). All three knobs are null; apparent
winners did not survive paired replication on an independent subset.}
\label{tab:ablation-hp}
\tabfont
\begin{tabular}{l l l}
\toprule
Knob & Values & Outcome \\
\midrule
$\alpha$ & 0, .25, .5, .75 & flat ($\pm 3$); $\alpha{=}1$ collapses (above) \\
learning rate & $3{\cdot}10^{-5}$, $10^{-4}$, $3{\cdot}10^{-4}$ & U-shaped $\pm 5$ $\Rightarrow$ noise \\
final temperature $T$ & .05, .15, .4 & flat \\
\bottomrule
\end{tabular}
\end{table}

\paragraph{Everything else is null or protective.}
The full sweep (Table~\ref{tab:ablation-hp}) found no improvement; the two
apparent single-subset winners ($\alpha{=}.25$, lr $3{\cdot}10^{-5}$)
reversed or split under paired replication. The same hyperparameter set is
used at both retentions --- the $\rho{=}50\%$ and $\rho{=}25\%$ checkpoints
differ in exactly one field, $K$. Finally, the transcript length filters
are protective rather than incidental: admitting very short responses
(fewer than 16 response tokens) adds examples whose per-example saliency is
averaged over too few tokens to be meaningful, and costs 21 F1 points on
the task those examples come from.

\section{Workflow rosters for the generalization ladder}
\label{app:rosters}

Tables~\ref{tab:roster-train}--\ref{tab:roster-l23} list every workflow used
in the generalization study of Section~\ref{sec:generalization}: the 32
training workflows, and the held-out sets at each rung. Scores are the
workflows' own validation scores under the dense backbone. AnsGen
abbreviates AnswerGenerate and ScEns the self-consistency ensemble operator.

\begin{table}[t]
\centering
\caption{Trained-on structures (32 workflows): only these train the
universal predictor.}
\label{tab:roster-train}
\tabfont\setlength{\tabcolsep}{3pt}
\begin{tabular}{lll}
\toprule
Task & Workflow & Operator sequence \\
\midrule
\multirow{12}{*}{HotpotQA}
 & debate\_1 r2      & Custom\,$\to$\,Custom\,$\to$\,Custom\,$\to$\,Custom \\
 & debate\_1 r3      & Custom\,$\to$\,Custom\,$\to$\,Custom\,$\to$\,Custom\,$\to$\,Custom \\
 & debate\_1 r5      & AnsGen\,$\to$\,AnsGen\,$\to$\,AnsGen\,$\to$\,ScEns\,$\to$\,Custom \\
 & debate\_1 r6      & AnsGen\,$\to$\,AnsGen\,$\to$\,Custom\,$\to$\,Custom \\
 & decompose\_1 r3   & Custom\,$\to$\,AnsGen\,$\to$\,Custom\,$\to$\,ScEns \\
 & decompose\_1 r6   & Custom\,$\to$\,AnsGen\,$\to$\,ScEns\,$\to$\,Custom \\
 & decompose\_1 r7   & Custom\,$\to$\,AnsGen\,$\to$\,Custom \\
 & decompose\_2 r3   & Custom\,$\to$\,AnsGen\,$\to$\,Custom \\
 & direct\_1 r1      & AnsGen \\
 & direct\_1 r3      & AnsGen\,$\to$\,ScEns \\
 & direct\_2 r3      & AnsGen\,$\to$\,ScEns \\
 & direct\_2 r5      & AnsGen\,$\to$\,ScEns\,$\to$\,Custom \\
\midrule
\multirow{13}{*}{HumanEval}
 & direct\_1 r1        & Custom \\
 & direct\_1 r2        & Custom\,$\to$\,Test \\
 & direct\_1 r4        & Custom\,$\to$\,Custom\,$\to$\,Custom \\
 & direct\_1 r7        & Custom\,$\to$\,Custom\,$\to$\,Custom\,$\to$\,Test \\
 & ensembletest\_1 r1  & Custom\,$\to$\,Test\,$\to$\,ScEns \\
 & ensembletest\_1 r2  & Custom\,$\to$\,Test\,$\to$\,ScEns\,$\to$\,Custom\,$\to$\,Test \\
 & ensembletest\_1 r6  & Custom\,$\to$\,Test\,$\to$\,ScEns \\
 & review\_1 r1        & Custom\,$\to$\,Custom\,$\to$\,Test \\
 & review\_2 r2        & Custom\,$\to$\,Custom\,$\to$\,Custom\,$\to$\,ScEns\,$\to$\,Test \\
 & review\_2 r6        & Custom\,$\to$\,Custom\,$\to$\,Custom\,$\to$\,Test \\
 & selfcons\_1 r1      & Custom\,$\to$\,ScEns \\
 & selfcons\_1 r3      & Custom\,$\to$\,ScEns\,$\to$\,Custom \\
 & selfcons\_2 r5      & Custom\,$\to$\,ScEns\,$\to$\,Custom\,$\to$\,Custom\,$\to$\,Test \\
\midrule
\multirow{7}{*}{MATH}
 & codefirst\_1 r1   & Programmer\,$\to$\,Custom \\
 & direct\_1 r1      & Custom \\
 & direct\_1 r2      & Custom\,$\to$\,ScEns \\
 & direct\_1 r3      & Custom \\
 & direct\_1 r4      & Custom\,$\to$\,ScEns\,$\to$\,Custom \\
 & direct\_2 r3      & Custom\,$\to$\,Custom \\
 & selfcons\_1 r2    & Custom\,$\to$\,ScEns\,$\to$\,Custom \\
\bottomrule
\end{tabular}
\end{table}

\begin{table}[t]
\centering
\caption{Level~0 held-out --- \textbf{unseen structures} (held-out rounds of
families the predictor \emph{did} train on).}
\label{tab:roster-l0}
\tabfont\setlength{\tabcolsep}{3pt}
\begin{tabular}{llcl}
\toprule
Task & Workflow & Score & Operator sequence \\
\midrule
\multirow{3}{*}{HotpotQA}
 & debate\_1 r4     & 0.719 & Custom\,$\to$\,Custom\,$\to$\,Custom\,$\to$\,Custom\,$\to$\,Custom \\
 & decompose\_2 r4  & 0.833 & Custom\,$\to$\,AnsGen\,$\to$\,AnsGen\,$\to$\,ScEns\,$\to$\,Custom \\
 & direct\_1 r4     & 0.453 & Custom\,$\to$\,AnsGen \\
\midrule
\multirow{2}{*}{HumanEval}
 & review\_1 r2     & 0.900 & Custom\,$\to$\,Custom\,$\to$\,ScEns\,$\to$\,Test \\
 & selfcons\_1 r6   & 0.960 & Custom\,$\to$\,Custom\,$\to$\,ScEns \\
\midrule
\multirow{2}{*}{MATH}
 & direct\_2 r2     & 0.833 & Custom\,$\to$\,ScEns \\
 & selfcons\_1 r1   & 0.767 & Custom\,$\to$\,ScEns \\
\bottomrule
\end{tabular}
\end{table}

\begin{table}[t]
\centering
\caption{Level~1 held-out --- \textbf{unseen families} (whole families never
seen in training: \emph{genverify} / \emph{gentestfix} / \emph{solveverify}).}
\label{tab:roster-l1}
\tabfont\setlength{\tabcolsep}{2pt}
\begin{tabular}{llcl}
\toprule
Task & Workflow & Score & Operator sequence \\
\midrule
\multirow{4}{*}{HotpotQA}
 & genverify\_1 r1  & 0.713 & AnsGen\,$\to$\,Custom\,$\to$\,Custom \\
 & genverify\_2 r3  & 0.793 & AnsGen\,$\to$\,ScEns\,$\to$\,Custom\,$\to$\,Custom \\
 & genverify\_2 r4  & 0.790 & AnsGen\,$\to$\,Custom\,$\to$\,Custom\,$\to$\,Custom \\
 & genverify\_2 r5  & 0.713 & AnsGen\,$\to$\,AnsGen\,$\to$\,AnsGen\,$\to$\,ScEns\,$\to$\,Custom\,$\to$\,Custom \\
\midrule
\multirow{3}{*}{HumanEval}
 & gentestfix\_1 r1 & 0.920 & Custom\,$\to$\,Test\,$\to$\,Custom\,$\to$\,Test \\
 & gentestfix\_1 r2 & 0.940 & Custom\,$\to$\,Custom\,$\to$\,Test\,$\to$\,Custom\,$\to$\,Test \\
 & gentestfix\_1 r3 & 0.940 & Custom\,$\to$\,Custom\,$\to$\,ScEns\,$\to$\,Test\,$\to$\,Custom\,$\to$\,Test \\
\midrule
\multirow{3}{*}{MATH}
 & solveverify\_1 r1 & 0.767 & Custom\,$\to$\,Programmer\,$\to$\,Custom \\
 & solveverify\_1 r2 & 0.900 & Custom\,$\to$\,Programmer\,$\to$\,Custom \\
 & solveverify\_1 r3 & 0.767 & Custom\,$\to$\,ScEns\,$\to$\,Programmer\,$\to$\,Custom \\
\bottomrule
\end{tabular}
\end{table}

\begin{table}[t]
\centering
\caption{Levels~2--3 held-out --- \textbf{unseen tasks}: MBPP = unseen code
benchmark, MMLU = unseen knowledge domain.}
\label{tab:roster-l23}
\tabfont\setlength{\tabcolsep}{2pt}
\begin{tabular}{llcl}
\toprule
Level / Task & Workflow & Score & Operator sequence \\
\midrule
\multirow{3}{*}{L2 $\cdot$ MBPP (code)}
 & MBPP\_1 r7   & 0.567 & CodeGen\,$\to$\,Custom\,$\to$\,Test\,$\to$\,Custom \\
 & MBPP\_1 r9   & 0.700 & CodeGen\,$\to$\,Test\,$\to$\,Custom \\
 & MBPP\_1 r14  & 0.700 & CodeGen\,$\to$\,ScEns\,$\to$\,Test\,$\to$\,Custom \\
\midrule
\multirow{3}{*}{L3 $\cdot$ MMLU (knowledge)}
 & MMLU\_1 r1   & 0.867 & Custom \\
 & MMLU\_1 r2   & 0.567 & AnsGen\,$\to$\,Custom\,$\to$\,Custom\,$\to$\,ScEns \\
 & MMLU\_1 r3   & 0.867 & AnsGen\,$\to$\,Custom\,$\to$\,Custom\,$\to$\,Custom\,$\to$\,ScEns \\
\bottomrule
\end{tabular}
\end{table}

\section{Per-task efficiency results}
\label{app:efficiency-full}

Table~\ref{tab:efficiency-full} gives the per-task breakdown behind the
averaged efficiency numbers of Table~\ref{tab:efficiency}, with seed std.

\begin{table}[t]
\centering
\caption{Per-task efficiency at $\rho{=}50\%$: mean $\pm$ std over 3 random
50-problem seeds, sequential contention-free runs.}
\label{tab:efficiency-full}
\tabfont
\begin{tabular}{llccc}
\toprule
task & policy & accuracy & tokens/prob & generation (s) \\
\midrule
\multirow{4}{*}{HotpotQA}
 & dense & .71$\pm$.03 & 172$\pm$8   & 412$\pm$70 \\
 & DEP   & \textbf{.75$\pm$.09} & 181$\pm$8   & \textbf{405$\pm$42} \\
 & REAP  & .36$\pm$.06 & 224$\pm$24  & 780$\pm$96 \\
 & freq  & .46$\pm$.08 & 477$\pm$28  & 894$\pm$26 \\
\midrule
\multirow{4}{*}{HumanEval+}
 & dense & .94$\pm$.03 & 1156$\pm$72 & 2548$\pm$266 \\
 & DEP   & .90$\pm$.03 & 1120$\pm$42 & \textbf{2391$\pm$130} \\
 & REAP  & .79$\pm$.02 & 1272$\pm$92 & 3113$\pm$318 \\
 & freq  & .00$\pm$.00 & 3352$\pm$39 & 4746$\pm$15 \\
\midrule
\multirow{4}{*}{MATH}
 & dense & .69$\pm$.18 & 4387$\pm$114 & 6757$\pm$298 \\
 & DEP   & .64$\pm$.13 & 4316$\pm$247 & 7030$\pm$346 \\
 & REAP  & .45$\pm$.17 & 4734$\pm$259 & 7516$\pm$178 \\
 & freq  & .36$\pm$.04 & 6266$\pm$43  & 10730$\pm$179 \\
\bottomrule
\end{tabular}
\end{table}

\section{Per-task results on a different MoE architecture}
\label{app:arch-full}

Table~\ref{tab:arch-full} gives the per-task breakdown behind the averaged
DeepSeek-V2.5 results of Table~\ref{tab:arch}.

\begin{table}[t]
\centering
\caption{DeepSeek-V2.5-236B (59 MoE layers, 160 routed experts, top-6, plus
2 shared experts that are always active and never pruned). Values are \%
($n{=}50$ problems per task; pass@1, HotpotQA token-F1, MATH/GSM8K exact
match); Dense and Direct are unpruned references. \textbf{Bold} = best
pruned method per task.}
\label{tab:arch-full}
\tabfont
\setlength{\tabcolsep}{3pt}
\begin{tabular}{l ccccc | c}
\toprule
\textbf{Method} & MATH & GSM8K & HumanEval+ & MBPP+ & HotpotQA & \textit{Mean} \\
\midrule
Dense (workflow)     & 52.0 & 84.0 & 72.0 & 86.0 & 59.7 & 70.7 \\
Direct (single call) & 50.0 & 90.0 & 76.0 & 84.0 & 56.9 & 71.4 \\
\midrule
\multicolumn{7}{l}{\emph{50\% retention (keep 80/160)}} \\
Freq.
 & 33.0 & 71.0 & 17.0 & 5.0 & \textbf{39.2} & 33.0 \\
REAP
 & 32.0 & 68.0 & 38.0 & 56.0 & 38.5 & 46.5 \\
\textbf{DEP}
 & \textbf{34.0} & \textbf{74.0} & \textbf{46.0} & \textbf{66.0} & 38.7 & \textbf{51.7} \\
\midrule
\multicolumn{7}{l}{\emph{25\% retention (keep 40/160)}} \\
Freq.
 & 1.0 & 2.0 & 0.0 & 0.0 & 1.0 & 0.8 \\
REAP
 & 2.0 & 2.0 & 0.0 & 2.0 & 1.7 & 1.5 \\
\textbf{DEP}
 & \textbf{4.0} & \textbf{4.0} & \textbf{2.0} & \textbf{12.0} & \textbf{12.5} & \textbf{6.9} \\
\bottomrule
\end{tabular}
\end{table}

\section{Token waste and generation collapse}
\label{app:collapse}

Table~\ref{tab:tokens} traces where the static masks' extra wall time comes
from, on HotpotQA at $\rho{=}50\%$ ($N{=}50$): they emit far more tokens per
problem, and the excess concentrates in \emph{max-token explosions} ---
generations that collapse into repetition loops and run to the cap. Dense and
DEP never explodes; each static mask does, on different inputs.
Table~\ref{tab:collapse} shows one example of each.

\begin{table}[t]
\centering
\caption{Token statistics of the final-answer agent (HotpotQA, $\rho{=}50\%$,
$N{=}50$). Exploded = generations $\ge$100 tokens; ms/token reflects batch
padding, not per-token compute.}
\label{tab:tokens}
\tabfont
\begin{tabular}{l rrrrrr}
\toprule
Policy & Accuracy (F1) & Tokens/prob & ms/token & Median tok & Max tok & \#\,exploded \\
\midrule
dense         & 0.73 & 180 & 47.3 & 4 & 21  & 0 \\
\textbf{DEP} & 0.70 & \textbf{188} & 45.5 & 4 & 11 & 0 \\
REAP          & 0.36 & 259 & 69.9 & 7 & 128 & 3 \\
freq          & 0.33 & 510 & 38.9 & 6 & 128 & 4 \\
\bottomrule
\end{tabular}
\end{table}

\begin{table}[t]
\centering
\caption{Generation collapse under static masks: the same input, all four
policies. Top: REAP collapses; bottom: freq collapses.}
\label{tab:collapse}
\tabfont
\begin{tabular}{l p{6.6cm} r r}
\toprule
Policy & Output & Tok & Score \\
\midrule
\multicolumn{4}{l}{\emph{Q: ``The saint in `A Time for Miracles' was born on what date?''}} \\
dense & \texttt{August 28, 1774} & 10 & 1.00 \\
DEP   & \texttt{August 28, 1774} & 10 & 1.00 \\
REAP  & \texttt{1774.1755-5.1815.1755-5.1815\dots} \textit{(repeats to max)} & 128 & 0.00 \\
freq  & \texttt{**Answer:**} \textit{(empty)} & 3 & 0.00 \\
\midrule
\multicolumn{4}{l}{\emph{Q: ``Gail Matthius co-anchored Weekend Update with the actor who}} \\
\multicolumn{4}{l}{\emph{played the villain Nicholas Andre in what movie?''}} \\
dense & \texttt{Dumb and Dumber} & 5 & 1.00 \\
DEP   & \texttt{"Dumb and Dumber"} & 6 & 1.00 \\
REAP  & \texttt{Dumb and Dumber (1988)} & 11 & 0.86 \\
freq  & \texttt{"D" in "D" in "D" in \dots} \textit{(repeats to max)} & 128 & 0.00 \\
\bottomrule
\end{tabular}
\end{table}

\section{Limitations}
\label{app:limitations}

This section expands the summary in Section~\ref{sec:conclusion}.

\paragraph{Training cost and its scope.}
DEP needs transcripts of the dense backbone running the training workflows,
and one predictor per retention level --- our $\rho{=}50\%$ and
$\rho{=}25\%$ checkpoints differ in exactly one field, $K$. The cost is
small (roughly 3 GPU-hours per level, Appendix~\ref{app:setup}) and is paid
once rather than per configuration, but it is a step static calibration does
not require, and serving a new retention level means training a new
predictor.

\paragraph{The training recipe is heuristic and the gap to dense is open.}
Pruned accuracy still trails the dense workflow at both retentions: we claim
graceful degradation under compression, not free compression. We do not
believe the current recipe is near the ceiling. Its objective mixes a
per-example saliency target with a self-distillation term in a fixed
proportion, and the mask keeps a uniform top-$K$ in every layer; both are
heuristic choices. The sweeps in Appendix~\ref{app:ablations} explore only a
neighbourhood of that design --- mixing weight, learning rate, annealing
temperature and the choice of saliency criterion --- and leave structurally
different alternatives untouched, such as an objective that scores a whole
candidate mask rather than experts individually, or a global budget
allocated unevenly across layers in place of a fixed $K$ per layer. We
expect a meaningful part of the remaining gap to dense to be recoverable
along those lines.

\paragraph{Fine-grained expert pools are assumed, and few architectures are tested.}
Masking requires $K \ge k$, so retention below $k/E$ cannot be expressed at
all, and just above it the router has little freedom left inside the kept
set. Every backbone we study has $E \ge 128$ routed experts; on
coarse-grained MoEs with few experts per layer this constraint binds quickly
and there are correspondingly few distinct masks to choose among, so we
expect the benefit of per-request specialization to shrink as expert
granularity coarsens. More broadly, our architectural coverage is thin: the
backbones we evaluate span two model families, so architecture independence
here is supported by example rather than established, and coarse-grained
designs in particular remain untested.

\paragraph{The workload must actually be heterogeneous.}
DEP's advantage comes from agents that need different experts. A workflow
whose nodes are near-identical gives the predictor little to specialize
over, and DEP should then behave much like a static mask fitted to that one
configuration.

\paragraph{Scope of the evidence.}
Our multi-agent workloads are automatically discovered workflows, and our
agentic-coding benchmark is run oracle-assisted; neither is a production
deployment, whose prompts, tool use, and traffic mix could differ in ways we
cannot anticipate here. The largest backbone we study is 480B, so
frontier-scale behavior is extrapolation. Validating DEP on production
multi-agent systems and at frontier scale is the most useful next test.

\paragraph{Measurement caveats.}
Several evaluations use small per-cell sample sizes --- stated in
Section~\ref{sec:setup} and in each table caption --- so individual cells
carry wide intervals and small per-cell differences should not be read as an
ordering. MATH is subset-sensitive in our harness --- dense accuracy varies
substantially between random draws of the same pool --- so MATH numbers are
interpretable only against a dense reference measured on the same problems.
The mask-overlap analysis of Section~\ref{sec:heterogeneity} is measured on
hand-crafted agent teams rather than the discovered workflows of
Table~\ref{tab:main}. The $\Delta$-load column of
Table~\ref{tab:efficiency} is an analytic worst case rather than a measured
swap latency, and the prefetching and agent-batching that would reduce it in
a production engine are described but not benchmarked end to end.

\end{document}